\documentclass[11pt]{article}

\usepackage[margin=1in]{geometry}

\usepackage{tech_report}

\usepackage[T1]{fontenc}
\usepackage[utf8]{inputenc}
\usepackage[scaled]{helvet}

\usepackage[table]{xcolor}
\usepackage{titlesec}
\usepackage{sectsty}
\usepackage{fancyhdr}
\usepackage{graphicx}
\usepackage{float}
\usepackage{booktabs,makecell}
\usepackage{wrapfig}
\usepackage[most]{tcolorbox}
\usepackage{hyperref}
\usepackage{enumitem}
\usepackage{ragged2e}
\usepackage{twemojis}
\usepackage{multirow}
\usepackage{times}
\usepackage{latexsym}
\usepackage{csquotes}
\usepackage{tabularray}
\usepackage{amsmath}
\usepackage{arydshln}
\usepackage{tikz}
\usepackage{pgfplots}
\pgfplotsset{compat=1.18}
\usepackage{subfigure}
\definecolor{darkyellow}{RGB}{251,188,4}
\definecolor{darkgreen}{RGB}{52,168,83}
\definecolor{lightblue}{RGB}{66,133,244}
\definecolor{acqua}{RGB}{70,189,198}

\tcbuselibrary{breakable}

\def\LRONE{$lr_{\text{S1}}$}
\def\LRTWO{$lr_{\text{S2}}$}
\def\PASR{$p_{\text{ASR}}$}

\newcommand{\github}{\includegraphics[width=12px]{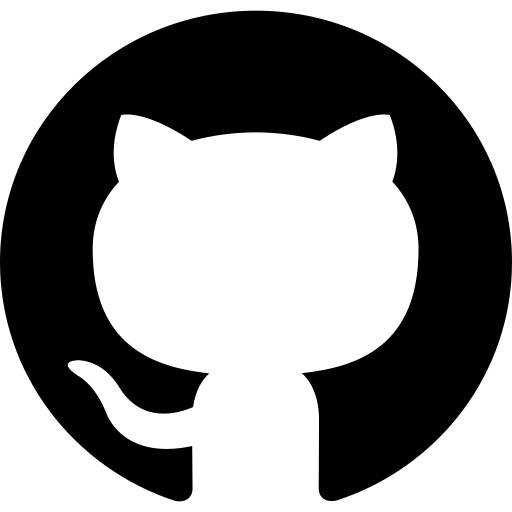}}
\newcommand{\huggingface}{\includegraphics[width=12px]{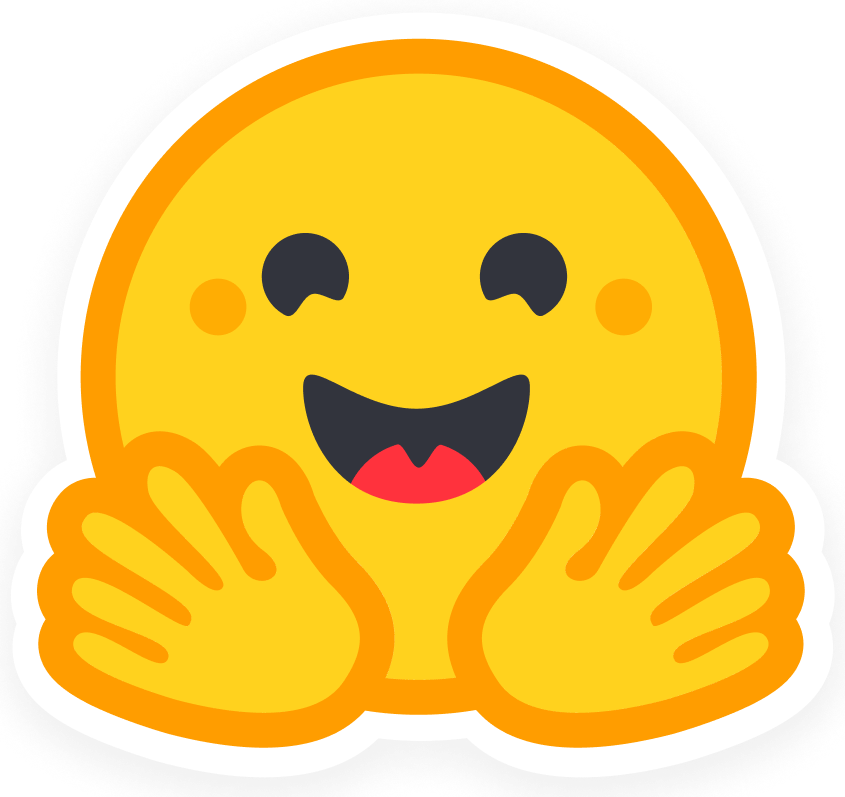}}
\newcommand{\fama}{\includegraphics[width=110px]{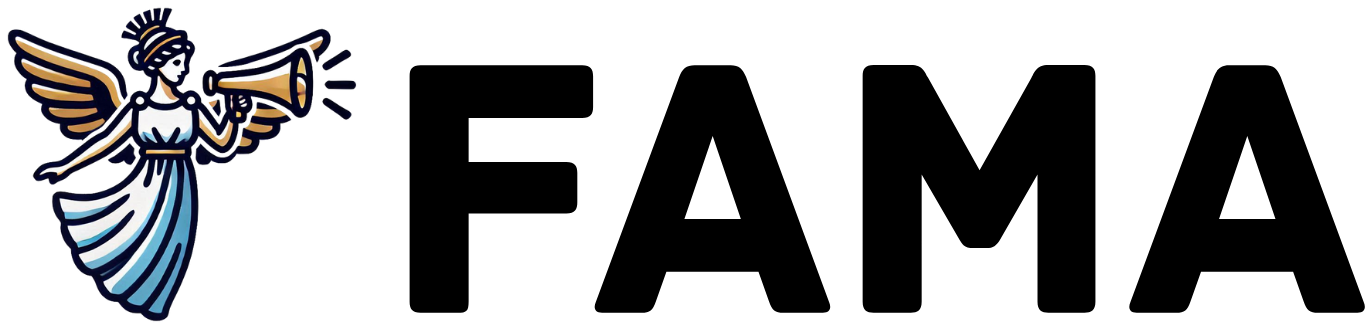}}

\definecolor{tuluBlue}{HTML}{2063BA}
\definecolor{tuluLightBlue}{HTML}{74B0FF}
\definecolor{tuluGray}{HTML}{333333}

\newcommand{\coaut}{\scalebox{1.25}{\twemoji{speech balloon}}}

\sectionfont{\color{tuluBlue}}
\subsectionfont{\color{tuluBlue}}
\subsubsectionfont{\color{tuluBlue}}

\titleformat{\section}
  {\Large\bfseries\color{tuluBlue}}{\thesection}{1em}{}
\titleformat{\subsection}
  {\large\bfseries\color{tuluBlue}}{\thesubsection}{1em}{}
\titleformat{\subsubsection}
  {\normalsize\bfseries\color{tuluBlue}}{\thesubsubsection}{1em}{}

\hypersetup{
    colorlinks=true,
    linkcolor=tuluBlue,
    urlcolor=tuluBlue,
    citecolor=tuluBlue
}

\title{\Huge\textbf{\color{tuluBlue}\fama{}: The First Large-Scale Open-Science Speech Foundation Model for English and Italian}}
\author{Sara Papi\textsuperscript{\coaut}, Marco Gaido\textsuperscript{\coaut}, Luisa Bentivogli, Alessio Brutti, Mauro Cettolo, \\ Roberto Gretter, Marco Matassoni, Mohamed Nabih, Matteo Negri \\ 
\textbf{Fondazione Bruno Kessler (FBK), Italy} \\
\texttt{\{spapi,mgaido,bentivo,brutti,cettolo}\\
\qquad\texttt{gretter,matasso,mnabih,negri\}@fbk.eu}\\
{\footnotesize\coaut{} \textit{denotes equal contribution}}}
\date{}

\begin{document}
\maketitle

\vspace{-2em}
\begin{flushleft}
\quad\huggingface\textbf{FAMA-medium (878M):} \url{https://hf.co/FBK-MT/fama-medium} \\
\quad\huggingface\textbf{FAMA-small (479M):} \url{https://hf.co/FBK-MT/fama-small} \\
\quad\huggingface\textbf{FAMA Data:} \url{https://hf.co/datasets/FBK-MT/fama-data} \\
\quad\github\textbf{FAMA Code:} \url{https://github.com/hlt-mt/FBK-fairseq} \\
\end{flushleft}

\vspace{0em}

\begin{tcolorbox}[
  colback=tuluGray!10,
  colframe=white,
  boxrule=0pt,
  enhanced jigsaw,
  left=12pt,
  right=12pt,
  top=8pt,
  bottom=8pt,
  before skip=12pt,
  after skip=12pt,
]

\begin{minipage}{0.75\linewidth}
{\Large\textbf{Abstract}} \vspace{0.2cm} \\
The development of speech foundation models (SFMs) like Whisper and SeamlessM4T has significantly advanced the field of speech processing. However, their closed nature--with inaccessible training data and code--poses major reproducibility and fair evaluation challenges. While other domains have made substantial progress toward open science by developing fully transparent models trained on open-source (OS) code and data, similar efforts in speech processing remain limited. To fill this gap, we introduce FAMA, the first family of open science SFMs for English and Italian, trained on 150k+ hours of OS speech data. Moreover, we present a new dataset containing 16k hours of cleaned and pseudo-labeled speech for both languages. Results show that FAMA achieves competitive performance compared to existing SFMs while being up to 8 times faster. All artifacts, including codebase, datasets, and models, are released under OS-compliant licenses, promoting openness in speech technology research.
\end{minipage}
\hfill
\begin{minipage}{0.20\linewidth}
  \includegraphics[width=\linewidth]{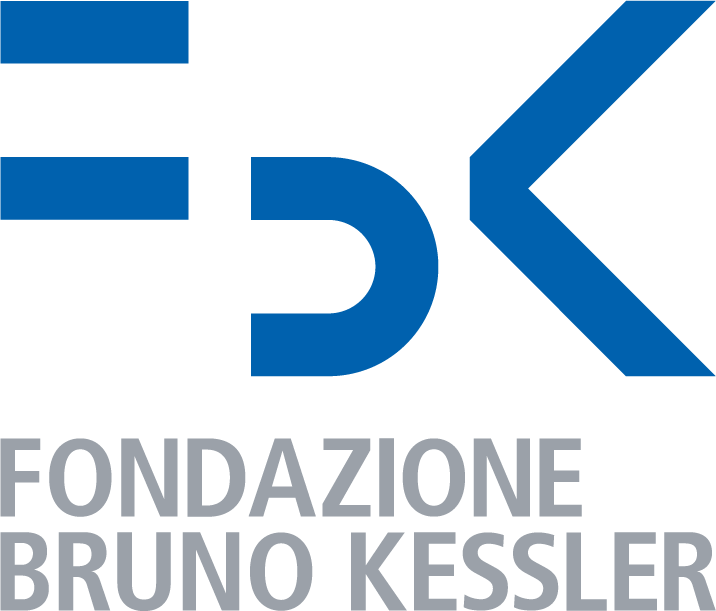} 
\end{minipage}

\end{tcolorbox}

\newpage

\tableofcontents

\newpage

\section{Introduction}

The development of speech foundation models (SFMs) has significantly advanced speech processing in the last few years, particularly in areas such as automatic speech recognition (ASR) and speech translation (ST). 
Popular SFMs such as OpenAI Whisper \citep{radford2023robust} and Meta SeamlessM4T \citep{barrault2023seamlessm4t} have been released to the public in various sizes and with extensive language coverage. However, these models completely lack comprehensive accessibility to their training codebases and datasets, hindering their reproducibility and raising concerns about potential data contamination \citep{dong-etal-2024-generalization}, thereby complicating fair evaluation.

In other domains, multiple efforts towards building models that are more accessible, reproducible, and free from proprietary constraints have been made \citep{workshop2022bloom,pmlr-v202-biderman23a,liu2023llm360,sun-etal-2024-openomni,deitke2024molmo,dai2024nvlm,martins2024eurollm}. For instance, the OLMO project \citep{groeneveld-etal-2024-olmo} has demonstrated the feasibility of training large language models (LLMs) using only open-source (OS) data \citep{soldaini-etal-2024-dolma}, realizing an \textit{open-science}\footnote{\textit{Open science} involves ensuring transparency and accessibility at all stages of the scientific process \citep{VICENTESAEZ2018428}, including publishing OS research papers, data, code, and any information needed to replicate the research.} system \citep{white2024model} for text processing. However, such comprehensive approaches are still lacking in the field of speech processing.

Recent works towards this direction
are represented by OWSM \citep{10389676} and its subsequent versions \citep{peng24b_interspeech}.
OWSM, whose model weights and the codebase used for the training are released open source, reproduces a Whisper-style training using publicly available data.
Despite representing a valuable initiative toward building an open-science system, there is still a step missing for creating the first SFM of this kind: leveraging only
data that is not only publicly available but also released under an OS-compliant license \citep{gaido-etal-2024-mosel}. Such effort would allow users complete access and control over the data used at every stage of the scientific process, promoting reproducibility \citep{belz-etal-2023-non}, fair evaluation \citep{balloccu-etal-2024-leak}, and the ability to build upon prior research without any barriers \citep{chesbrough2015open}.
Besides transparency and collaboration, these efforts also foster users' trust by ensuring that data is not leveraged to build tools that can be used under conditions/purposes (e.g., commercial) for which the data was not intended \citep{white2024model}.

To fill this gap, we create \textbf{FAMA},\footnote{\textit{Fama} (from the Latin \enquote{fari} meaning \enquote{to speak}) is the personification of the public voice in Roman mythology.} the first family of large-scale open-science SFMs for English and Italian trained on over 150k hours of exclusively OS-compliant speech data. 
We leverage both already available OS datasets and create a new collection of ASR and ST psuedolabels for Italian and English comprising more than 16k hours of OS-compliant speech, along with automatically generated Italian and English translations for an additional 130k+ hours of speech.
We also detail training and evaluation procedures and provide full access to training data to have complete control of the model creation and avoid data contamination issues.
FAMA models achieve remarkable results, with up to 4.2 WER and 0.152 COMET improvement on average across languages compared to OWSM and remaining competitive in terms of ASR performance with the Whisper model family while being up to 8 times faster.
All the artifacts used for realizing FAMA models, including codebase, datasets, and models themself, are released under OS-compliant licenses, promoting a more responsible creation of models in our community.
Our approach would not only facilitate fair evaluation and comparison of SFMs but also encourage broader participation in speech technology development, leading to more inclusive and diverse applications.

\section{The FAMA Framework}
\subsection{Training and Evaluation Data}

In compliance with the open-science ideology, we train and test our models only on OS-compliant data. The training set comprises both already publicly available OS datasets, 
and new pseudolabels created for this work, whose list is presented in Table \ref{tab:public-train-data}.

\begin{table}[!ht]
    \centering
    \begin{tblr}{
        columns={colsep=2pt},
        rows={rowsep=0.5pt},
        row{3,5,7,9,11} = {bg=tuluGray!10},
      colspec={|X[1.3]|X[0.3,r]|X[0.3,r]|X[0.25,c]|}, row{1-2} = {c}, hlines,
    }
        \SetCell[r=2]{c}\textbf{Dataset} & \SetCell[c=2]{c}{\textbf{\#hours}} & & \SetCell[r=2]{c}\textbf{Label} \\
        & \color{tuluGray!75}\textit{en} & \color{tuluGray!75}\textit{it} & \\
        \hline
         CommonVoice v18 \citep{commonvoice:2020} & 1746 & 250 & G \\
         CoVoST2 \citep{wang21s_interspeech} & 420 & 28 & G \\
         FLEURS \citep{10023141} & 7 & 9 & G \\
         LibriSpeech \citep{7178964} & 358 & - & G \\
         MOSEL \citep{gaido-etal-2024-mosel} & 66,301 & 21,775 & \SetCell[r=1]{c}A \\
         MLS \citep{pratap20_interspeech} & 44,600 & 247 & G \\
         VoxPopuli-ASR \citep{wang-etal-2021-voxpopuli} & 519 & 74 & G \\
         YouTube-Commons (\textit{our paper}) & 14,200 & 1,828 & A \\
        {\SetCell[r=1]{c}\textit{\textbf{Total}}} & \textbf{128,152} & \textbf{24,211} & G+A \\
    \end{tblr}
    \caption{List of both publicly available training data and the data created in this paper for English (en) and Italian (it). \enquote{G} stands for gold labels while \enquote{A} for automatically generated labels (transcripts).}
    \label{tab:public-train-data}
\end{table}

To create the new pseudolabels, we leveraged the speech content of
YouTube-Commons,\footnote{\url{https://hf.co/datasets/PleIAs/YouTube-Commons}} a dataset collecting YouTube videos released with the permissive CC-BY 4.0 license. The videos are automatically converted into wav files with one channel and a sampling rate of 16k Hz. Then, the audio is cleaned from music and non-speech phenomena and segmented using silero \citep{silero}, a lightweight VAD having low computational requirements. Lastly, the audio is split using SHAS \citep{tsiamas22_interspeech} to obtain segments suitable for training of around 16 seconds on average. 
The resulting dataset contains automatic transcripts, which we created with Whisper \texttt{large-v3},\footnote{\url{https://hf.co/openai/whisper-large-v3}} for 14,200 hours of speech for English (\textit{en}) and 1,828 for Italian (\textit{it}).
Including publicly available data (113,951 hours for \textit{en}, and 22,383 hours for \textit{it}),
the final ASR training set comprises 128,152 hours of \textit{en} speech and 24,211 hours of \textit{it} speech, with a total of 152,363 hours of speech data, including 48,259 gold-labeled hours. 

\begin{table}[!ht]
    \centering
    \begin{tblr}{
        columns={colsep=2pt},
        rows={rowsep=0.5pt},
        row{3,5,7,9,11,13} = {bg=tuluGray!10},
      colspec={|X[1.3]|X[0.3,r]|X[0.3,r]|X[0.25,c]|}, row{1-2} = {c}, hlines,
    }
        \SetCell[r=2]{c}\textbf{Dataset} & \SetCell[c=2]{c}{\textbf{\#hours}} & & \SetCell[r=2]{c}\textbf{Label} \\
        & \color{tuluGray!75}\textit{en-it} & \color{tuluGray!75}\textit{it-en} & \\
        \hline
         CommonVoice v18 \citep{commonvoice:2020} & 1746 & 250 & A \\
         CoVoST2 \citep{wang21s_interspeech} - automatic labels & 420 & 28 & A \\
         LibriSpeech \citep{7178964} & 358 & - & A \\
         MOSEL \citep{gaido-etal-2024-mosel} & 66,301 & 21,775 & \SetCell[r=1]{c}A \\
         MLS \citep{pratap20_interspeech} & 44,600 & 247 & A \\
         VoxPopuli-ASR \citep{wang-etal-2021-voxpopuli} & 519 & 74 & A \\
         YouTube-Commons (\textit{our paper}) & 14,200 & 1,828 & A \\
         {\SetCell[r=1]{c}\textit{Total (A)}} & 128,144 & 24,202 & A \\
         {\SetCell[r=1]{c}\textit{Filtered (A)}} & 123,777 & 23,445 & A \\
        CoVoST2 \citep{wang21s_interspeech} - gold labels & 420 & 28 & G \\
         FLEURS \citep{10023141} & 7 & 9 & G \\
        {\SetCell[r=1]{c}\textit{\textbf{Total}}} & \textbf{124,204} & \textbf{23,482} & G+A \\
    \end{tblr}
    \caption{List of both publicly available training data and the data created in this paper for English-Italian (en-it) and Italian-English (it-en). \enquote{G} stands for gold labels while \enquote{A} for automatically generated labels (translations). }
    \label{tab:train-data-st}
\end{table}

Being composed of speech-transcript pairs, the data mentioned so far is suitable for ASR. For ST, instead, only CoVoST2 and FLEURS contain translations from and into \textit{en} and \textit{it}.
For this reason, we automatically translated the transcripts of all the speech data (including the original CoVoST2) with MADLAD-400 \texttt{3B-MT} \citep{kudugunta2023madlad}.\footnote{\url{https://hf.co/google/madlad400-3b-mt}} 
Following \citep{gaido-etal-2022-efficient,alam2024case}, we additionally filter out samples based on the ratio $r$ between the source and target text lengths (in characters) for each language pair based on their distribution (${r_{\text{min}}=0.75,r_{\text{max}}=1.45}$ for en-it, and ${r_{\text{min}}=0.65,r_{\text{max}}=1.35}$ for it-en), resulting into 3.41\% of data filtering for en-it and 3.12\% for it-en. The final training set (Table \ref{tab:train-data-st}) comprises the automatically translated speech data and the gold CoVoST2 and FLEURS datasets, resulting in a total of 147,686 hours for \textit{en-it} and \textit{it-en}.

For training, validation, and testing, we use gold-labeled benchmarks. ASR evaluation is conducted on CommonVoice, MLS, and VoxPopuli, with CommonVoice also serving as the validation set for both \textit{en} and \textit{it}. For translation, we use CoVoST2 for \textit{it-en} and FLEURS dev and test sets for \textit{en-it}.

\subsection{Model Architecture}

FAMA models are two-sized encoder-decoder architectures, \texttt{small} and \texttt{medium}. Both models are composed of a Conformer encoder \citep{gulati20_interspeech} and a Transformer decoder \citep{transformer}. FAMA \texttt{small} has 12 encoder layers and 6 decoder layers, while FAMA \texttt{medium} has 24 encoder layers and 12 decoder layers. 
Our decision to use an encoder twice as deep as the decoder--unlike Whisper and OWSM, which have an equal number of encoder and decoder layers--is driven by two key motivations: \textit{i)} since autoregressive models perform multiple decoder passes during output generation, a shallower decoder speeds up inference by making each pass faster, and \textit{ii)} since many approaches integrate SFMs with LLMs by leveraging the encoder \citep{gaido-etal-2024-speech}, a deeper encoder helps preserve more of the SFMs processing capabilities in such integrations.
Each layer has 16 attention heads, an embedding dimension of 1,024, and an FFN dimension of 4,096.


The Conformer encoder is preceded by two 1D convolutional layers with a stride of 2 and a kernel size of 5. The kernel size of the Conformer convolutional module is 31 for both the point- and depth-wise convolutions. The vocabulary is built using a SentencePiece unigram model \citep{kudo-richardson-2018-sentencepiece} with size 16,000 trained on \textit{en} and \textit{it} transcripts. Two extra tokens--\texttt{<lang:en>} and \texttt{<lang:it>}--are added to indicate whether the target text is in \textit{en} or \textit{it}. The input audio is represented by 80 Mel-filterbank features extracted every 10 ms with a window of 25 ms.

\subsection{Training and Evaluation Procedures}
\label{subsec:train-eval-procedure}

We train both models using a combination of three losses. 
 First, a label-smoothed cross-entropy loss ($\mathcal{L}_{\text{CE}}$) is applied to the decoder output, using the target text as the reference (transcripts for ASR and translations for ST). Second, a CTC loss \citep{ctc-2006} is computed using transcripts as reference ($\mathcal{L}_{\text{CTCsrc}}$) on the output of the 8\textsuperscript{th} encoder layer for \texttt{small} and the 16\textsuperscript{th} for \texttt{medium}. Third, a CTC loss on the final encoder output ($\mathcal{L}_{\text{CTCtgt}}$) is applied to predict the target text.
The final loss is the weighted sum of the above-mentioned losses:
\begin{equation*}
    \mathcal{L} = \lambda_1 \mathcal{L}_{\text{CE}} + \lambda_2 \mathcal{L}_{\text{CTCsrc}} + \lambda_3 \mathcal{L}_{\text{CTCtgt}}
\end{equation*}
where $\lambda_1,\lambda_2,\lambda_3=5.0,1.0,2.0$, and the label smoothing factor of the CE is $0.1$.

FAMA models are trained using a two-stage approach, where the model is pre-trained first on ASR data only (ASR pre-training) and then trained on both ASR and ST data (ASR+ST training).
Both training stages lasted 1M steps, corresponding to $\sim$6 epochs over the training data.

For the ASR pre-training, the learning rate (\LRONE) scheduler adopted to train the \texttt{small} model is the Noam scheduler \citep{transformer} with a peak of 2e-3 and 25,000 warm-up steps.
To cope with convergence issues similar to \citep{peng24b_interspeech}, for the \texttt{medium} model
we adopted a piece-wise warm-up on the Noam scheduler, with the learning rate first increasing linearly to 2e-5 for 25k steps and then to 2e-4 for an additional 25k steps, followed by the standard inverse square root function.
For the ASR+ST training, we sample the ASR target with probability \PASR{}=0.5 and use the ST target otherwise. Training settings are the same as for ASR pre-training, except for the learning rate that is set to a constant value \LRTWO=1e-4. Experiments on how 
\PASR{} and \LRTWO{} are determined for the \texttt{small} model are discussed in Section \ref{subsec:pre-train}. For the \texttt{medium} model, similarly to the first stage, the \LRTWO{} is scaled down by one order of magnitude compared to the \texttt{small} model, i.e., a constant value \LRTWO=1e-5 is used.

The optimizer is AdamW with momentum $\beta_1,\beta_2=0.9,0.98$, a weight decay of $0.001$, a dropout of $0.1$, and clip normalization of $10.0$.
We apply SpecAugment \citep{Park2019} during both ASR pre-training and ASR+ST training.
We use mini-batches of 10,000 tokens for FAMA \texttt{small} and 4,500 for FAMA \texttt{medium} with an update frequency of, respectively, 2 and 6 on 16 NVIDIA A100 GPUs (64GB RAM), save checkpoints every 1,000 steps and average the last 25 checkpoints to obtain the final model.

The inference is performed using a single NVIDIA A100 GPU with a batch size of 80,000 tokens. We use beam search with beam 5, unknown penalty of 10,000, and
no-repeat n-gram size of 5. Additionally, we report the results using the joint CTC rescoring \citep{yan-etal-2023-ctc}, leveraging the 
CTC on the encoder output with weight 0.2.
Both training and inference are done using the bug-free Conformer implementation \citep{papi-etal-2024-good} available in FBK-fairseq,\footnote{\url{https://github.com/hlt-mt/FBK-fairseq}} which is built upon fairseq-S2T \citep{wang2020fairseqs2t}. 
ASR performance is evaluated with word error rate (WER) using the jiWER library\footnote{\url{https://pypi.org/project/jiwer/}} with the text normalized using Whisper normalizer\footnote{\url{https://pypi.org/project/whisper-normalizer/}}.
ST performance is evaluated using COMET \citep{rei-etal-2020-comet} version 2.2.4, with the default \texttt{Unbabel/wmt22-comet-da}
model.

\subsection{Terms of Comparison}
\label{subsec:terms-of-comparison}

As a first term of comparison, we use Whisper \citep{radford2023robust} in both \texttt{medium}\footnote{\url{https://hf.co/openai/whisper-medium}} and \texttt{large-v3} configurations as the first is comparable with FAMA \texttt{medium} in terms of size and the second--trained on more than 4M hours---is the best performing model of the Whisper family.
The comparison is made for \textit{en} and \textit{it} ASR and \textit{it-en} ST, as Whisper does not cover the \textit{en-to-many} translation directions. 
Whisper models are released under Apache 2.0 license and, therefore, open weights.
For both ASR and ST, we also compare with SeamlessM4T 
\texttt{medium}\footnote{\url{https://hf.co/facebook/hf-seamless-m4t-medium}} and \texttt{v2-large}\footnote{\url{https://hf.co/facebook/seamless-m4t-v2-large}} covering ASR and both ST language directions \citep{barrault2023seamlessm4t}. The model is non-commercial and, therefore, not open.
We also compare with OWSM \texttt{v3.1 medium}\footnote{\url{https://hf.co/espnet/owsm_v3.1_ebf}}, the best performing model of the OWSM family, also covering both ASR and ST language directions and released open source \citep{peng24b_interspeech}.

To ensure a fair comparison, we 
perform the inference with HuggingFace transformers\footnote{\url{https://pypi.org/project/transformers/}} version 4.48.1 using the standard settings and beam search with beam 5, except for OWSM, which is not supported on HuggingFace, and for which the original ESPNet\footnote{\url{https://github.com/espnet/espnet/tree/master/egs2/owsm_v3.1/s2t1}} inference code is used with a beam size of 3.\footnote{We attempted to use a beam size of 5 but the model had out-of-memory issues even when reducing the batch size.} 

\section{Results}

\subsection{Pre-training and Catastrophic Forgetting}
\label{subsec:pre-train}

Catastrophic forgetting is a well-known problem in machine learning \citep{MCCLOSKEY1989109} that arises when a system is trained sequentially on multiple languages or tasks, leading to a degradation in performance on original domains or languages \citep{10.1145/3534678.3539169}. As we follow a two-stage approach, which is commonly employed in SFMs training \citep{radford2023robust}, we analyze the conditions in which this phenomenon arises during the ASR+ST training.

\pgfplotstableread[row sep=\\]{
steps   pplasr	pplst \\
10	1.815	7.53 \\
20	1.795	7.12 \\
30	1.795	6.705 \\
40	1.79	6.5 \\
50	1.785	6.255 \\
60	1.78	6.215 \\
70	1.77	6.22 \\
80	1.8	6.12 \\
90	1.775	6.07 \\
100	1.77	5.955 \\
}\highlrhighST

\pgfplotstableread[row sep=\\]{
steps   pplasr	pplst \\
10	1.62	8.27 \\
20	1.62	7.595 \\
30	1.615	7.15 \\
40	1.61	6.87 \\
50	1.61	6.7 \\
60	1.605	6.565 \\
70	1.605	6.465 \\
80	1.605	6.39 \\
90	1.605	6.29 \\
100	1.6	6.195 \\
110 1.605	6.14 \\
120 1.595	6.08 \\
130 1.595	6.055 \\
140 1.59	6.005 \\
150 1.59	5.98 \\
160 1.595	5.935 \\
170 1.59	5.92 \\
190 1.59	5.835 \\
200 1.585	5.805 \\
210 1.59	5.77 \\
220 1.59	5.745 \\
230 1.585	5.755 \\
240 1.59	5.715 \\
250 1.59	5.65 \\
260 1.585	5.7 \\
270 1.58	5.61 \\
280 1.585	5.645 \\
290 1.58	5.615 \\
300 1.58	5.61 \\
310 1.58	5.575 \\
320 1.585	5.565 \\
330 1.58	5.555 \\
340 1.58	5.53 \\
350 1.58	5.53 \\
360 1.58	5.49 \\
370 1.58	5.475 \\
380 1.58	5.485 \\
390 1.58	5.46 \\
400 1.58	5.46 \\
410 1.575	5.465 \\
420 1.58	5.425 \\
430 1.575	5.425 \\
440 1.58	5.405 \\
450 1.575	5.39 \\
460 1.58	5.39 \\
470 1.58	5.42 \\
480 1.57	5.35 \\
490 1.575	5.345 \\
500 1.58	5.35 \\
}\lowlrhighST

\pgfplotstableread[row sep=\\]{
steps   pplasr	pplst \\
10	1.72	8.12 \\
20	1.725	7.63 \\
30  1.705	7.14 \\
40  1.705	6.785 \\
50  1.7	6.675 \\
60  1.69	6.53 \\
70  1.69	6.625 \\
80  1.71	6.53 \\
90  1.685	6.36 \\
100 1.7	6.385 \\
}\highlrequiST

\pgfplotstableread[row sep=\\]{
steps   pplasr	pplst \\
10	1.55	8.755 \\
20	1.545	7.955 \\
30  1.555	7.53 \\
40  1.545	7.215 \\
50  1.54	7 \\
60  1.545	6.895 \\
70  1.54	6.795 \\
80  1.54	6.675 \\
90  1.54	6.585 \\
100 1.545	6.48 \\
110 1.54	6.405 \\
120 1.54	6.35 \\
130 1.535	6.35 \\
140 1.535	6.24 \\
150 1.535	6.25 \\
160 1.53	6.18 \\
170 1.53	6.13 \\
180 1.53	6.09 \\
190 1.53	6.065 \\
200 1.53	6.025 \\
210 1.53	6.005 \\
220 1.53	5.975 \\
230 1.525	5.975 \\
240 1.535	5.925 \\
250 1.53	5.89 \\
260 1.525	5.87 \\
270 1.525	5.815 \\
280 1.525	5.835 \\
290 1.53	5.805 \\
300 1.53	5.775 \\
310 1.53	5.74 \\
320 1.53	5.755 \\
330 1.53	5.74 \\
340 1.525	5.725 \\
350 1.52	5.72 \\
360 1.525	5.7 \\
370 1.52	5.69 \\
380 1.52	5.685 \\
390 1.52	5.645 \\
400 1.525	5.65 \\
410 1.525	5.645 \\
420 1.525	5.61 \\
430 1.525	5.64 \\
440 1.525	5.62 \\
450 1.52	5.62 \\
460 1.52	5.545 \\
470 1.525	5.57 \\
480 1.52	5.56 \\
490 1.525	5.525 \\
500 1.52	5.535 \\
}\lowlrequiST

\pgfplotstableread[row sep=\\]{
steps   pplasr \\
0   1.52 \\
1000    1.52 \\
}\pretraining

\begin{figure}[!ht]
\centering
\footnotesize
\subfigure[ASR]{
\begin{tikzpicture}
    \begin{axis}[
            ymajorgrids=true,
            xtick pos=left,
            ytick pos=left,
            minor y tick num=1,
            minor x tick num=0,
            ymin=1.5,
            ymax=1.825,
            xmin=10,
            xmax=500,
            ylabel=ppl, xlabel=steps (k),
            ylabel shift={-7pt},
            width=8.25cm,
            height=6cm,
            xtick=data,
            compat=newest,
            xtick={50,100,150,200,250,300,350,400,450,500},
            every axis plot/.append style={thick},
            legend to name=sharedlegend,
            legend style={at={(0.5,1.35)},anchor=north,legend columns=2},
        ]
        \addplot[color=lightblue, mark=] table[x=steps,y=pplasr]{\highlrhighST};
        \addplot[color=orange, mark=] table[x=steps,y=pplasr]{\lowlrhighST};
        \addplot[color=darkgreen, mark=] table[x=steps,y=pplasr]{\highlrequiST};
        \addplot[color=magenta, mark=] table[x=steps,y=pplasr]{\lowlrequiST};
        \addplot[dashed, color=black] table[x=steps,y=pplasr]{\pretraining};
        \addlegendentry{\LRTWO=1e-3, \PASR=0.2}
        \addlegendentry{\LRTWO=1e-4, \PASR=0.2}
        \addlegendentry{\LRTWO=1e-3, \PASR=0.5}
        \addlegendentry{\LRTWO=1e-4, \PASR=0.5}
    \end{axis}
\end{tikzpicture}}
\subfigure[ST]{
\begin{tikzpicture}
    \begin{axis}[
            ymajorgrids=true,
            xtick pos=left,
            ytick pos=left,
            minor y tick num=1,
            minor x tick num=0,
            ymin=5,
            ymax=9,
            xmin=10,
            xmax=500,
            ylabel=ppl, xlabel=steps (k),
            ylabel shift={-4pt},
            width=8.25cm,
            height=6cm,
            xtick=data,
            compat=newest,
            xtick={50,100,150,200,250,300,350,400,450,500},
            every axis plot/.append style={thick},
            legend style={at={(0.5,1.3)},anchor=north,legend columns=4},
        ]
        \addplot[color=lightblue, mark=] table[x=steps,y=pplst]{\highlrhighST};
        \addplot[color=orange, mark=] table[x=steps,y=pplst]{\lowlrhighST};
        \addplot[color=darkgreen, mark=] table[x=steps,y=pplst]{\highlrequiST};
        \addplot[color=magenta, mark=] table[x=steps,y=pplst]{\lowlrequiST};
    \end{axis}
\end{tikzpicture}
}
\vspace{0.5em}
\centering
\ref{sharedlegend}
\caption{Average ASR and ST perplexity (ppl) on both English and Italian up to 500k steps of the training. Due to the evident worse results achieved by using a $lr$ of 1e-3, we stopped the training curves after 100k steps. The black dashed line is the ppl of the ASR model from which the training is started.}
\label{fig:ppl}
\end{figure}
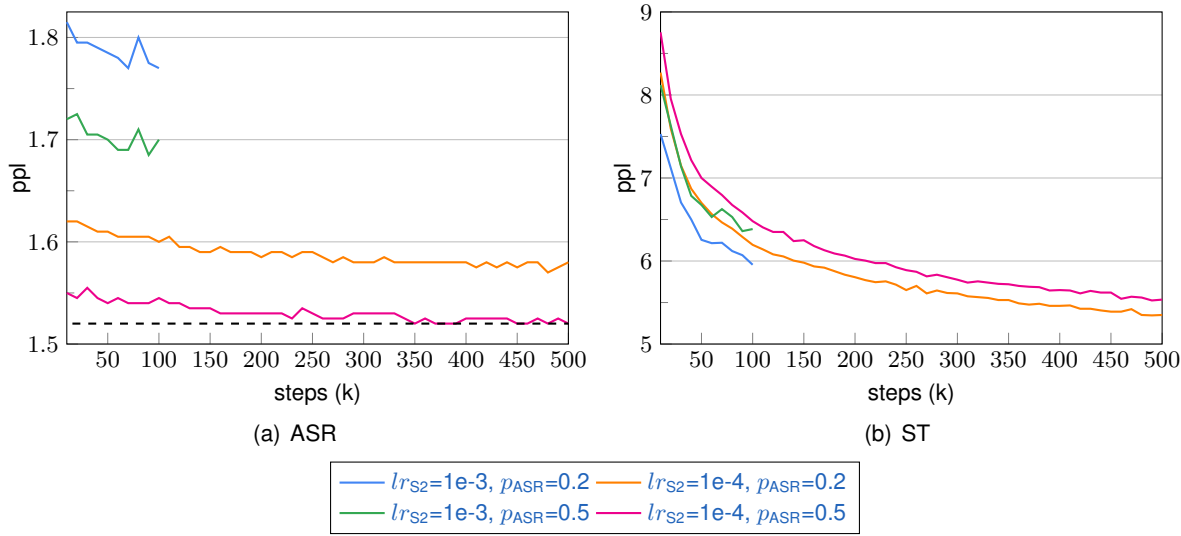

Figure \ref{fig:ppl} shows the perplexity (ppl) behavior during the first 100/500k steps of the FAMA \texttt{small} model training on the validation sets. We present the results of different systems obtained by varying both the learning rate \LRTWO{} and the sampling probability \PASR{} discussed in Section \ref{subsec:train-eval-procedure}. Lower values of \LRTWO{} (e.g., 1e-5) lead to worse performance and are not included in the results. Since the computational budget for our experiments is limited, we analyze two cases for the sampling probability: 1) \PASR=0.5 to obtain a system equally trained on both ASR and ST tasks, and 2) \PASR=0.2 to obtain a system trained more on the unseen task during pre-training i.e., the ST task.

As we can see from the curves, a \LRTWO{} of 1e-3 seems to be too high for maintaining good ASR performance while learning a new task (ST). Both in the case in which the ST training is more boosted (\PASR=0.2) and in the case in which ASR and ST training is balanced (\PASR=0.5), we notice a significant increase in the ASR ppl of up to 0.25 that corresponds to a drop in performance of 3-4 WER on both languages -- which, moreover, it is not recovered later on in the training. Therefore, to avoid catastrophic forgetting arising just in the first steps, we exclude \LRTWO=1e-3 and use 1e-4 for the two-stage training. Regarding the ASR sampling, we look at the behavior of the curves for 500k steps (half of the second-stage training) and notice that the ASR ppl curve with \PASR=0.5 slowly approaches the original model ppl value while the one with \PASR=0.2, despite improving, is not able to approach the original ppl value.
This is counterbalanced by a lower (hence, better) ppl of the \PASR=0.2 curve on ST compared to that of the \PASR=0.5 curve. However, this difference, which is about $\sim$0.2 ppl, is not reflected in the ST performance, which only improves by 0.005 COMET points on average. Instead, the difference in terms of WER is significant, with a quality drop of $\sim$0.8 WER across en and it.
As a result, we conclude that we avoid catastrophic forgetting in the two-stage training only by evenly sampling the ASR and ST tasks during the second step. 

\begin{table*}[!ht]
\footnotesize
  \centering
  \begin{tabular}{l|c!{\vrule width 1.2pt}cc|cc|cc|cc!{\vrule width 1.2pt}c|c}
    \Xhline{1.2pt}
    \multirow{3}{*}{\color{tuluBlue}\textbf{Model}} & \multirow{3}{*}{\color{tuluBlue}\textbf{\#params}} & \multicolumn{8}{c!{\vrule width 1.2pt}}{\color{tuluBlue}\textbf{ASR (WER $\downarrow$)}} & \multicolumn{2}{c}{\color{tuluBlue}\textbf{ST (COMET $\uparrow$)}} \\
    \cline{3-12} 
     & & \multicolumn{2}{c|}{\textbf{CV}} & \multicolumn{2}{c|}{\textbf{MLS}} & \multicolumn{2}{c|}{\textbf{VP}} & \multicolumn{2}{c!{\vrule width 1.2pt}}{\color{tuluBlue!85!black}\textbf{\underline{AVG}}} & \multicolumn{1}{c|}{\textbf{CVST2}} & \multicolumn{1}{c}{\textbf{FLRS}} \\
     \cline{3-12}
     & & \color{tuluGray!75}\textit{en} & \color{tuluGray!75}\textit{it} &  \color{tuluGray!75}\textit{en} & \color{tuluGray!75}\textit{it} & \color{tuluGray!75}\textit{en} & \color{tuluGray!75}\textit{it} & \color{tuluGray!75}\textit{en} & \color{tuluGray!75}\textit{it} & \color{tuluGray!75}\textit{it-en} & \color{tuluGray!75}\textit{en-it} \\
    \hline
    Whisper \texttt{medium} & 769M & 14.5 & 10.4 & 14.2 & 15.9 & 8.1 & 26.8 & 12.3 & 17.7 & 0.801 & - \\
    Whisper \texttt{large-v3}& 1550M & 11.2 & 6.5 & \textbf{5.0} & 8.8 & 7.1 & 18.8 & 7.8 & 11.4 & 0.825 & - \\
    OWSM v3.1 \texttt{medium} & 1020M & 11.9 & 12.5 & 6.6 & 19.3 & 8.4 & 24.0 & 9.0 & 18.6 & 0.636 & 0.337 \\ %
    SeamlessM4T \texttt{medium} & 1200M & 10.7 & 7.8 & 8.8 & 11.3 & 10.2 & 18.2 & 9.9 & 12.4 & 0.831 & 0.820 \\    
    SeamlessM4T \texttt{v2-large} & 2300M & \textbf{7.7} & \textbf{5.0} & 6.4 & \textbf{8.5} & \textbf{6.9} & 16.6 & \textbf{7.0} & \textbf{10.0} & \textbf{0.852} & \textbf{0.855} \\
    \hline
    \rowcolor{tuluGray!10}
    FAMA-ASR \texttt{small} & \multirow{2}{*}{475M} & 13.8 & 8.9 & 5.8 & 12.6 & 7.2 & 15.7 & 8.9 & 12.4 & - & - \\
    \textcolor{tuluBlue}{\quad \textit{+ joint CTC rescoring}} & & 13.9 & 8.9 & 5.8 & 12.4 & 7.0 & \textbf{14.6} & 8.9 & 12.0 & - & - \\
    \hdashline
    \rowcolor{tuluGray!10}
    FAMA-ASR \texttt{medium} & \multirow{2}{*}{878M} & 11.7 & 7.1 & 5.1 & 12.2 & 7.0 & 15.9 & 7.9 & 11.7 & - & - \\
    \textcolor{tuluBlue}{\quad \textit{+ joint CTC rescoring}} &  & 11.7 & 7.0 & 5.1 & 12.2 & 7.0 & \textbf{14.6} & 7.9 & 11.3 & - & - \\
    \hline
    \rowcolor{tuluLightBlue!30}
    FAMA \texttt{small} & \multirow{2}{*}{475M} & 13.7 & 8.6 & 5.8 & 12.8 & 7.3 & 15.6 & 8.9 & 12.3 & 0.774 & 0.807 \\
     \textcolor{tuluBlue}{\quad \textit{+ joint CTC rescoring}} & & 13.6 & 8.5 & 5.8 & 12.8 & 7.2 & 14.8 & 8.9 & 12.0 & 0.777 & 0.804 \\
  \hdashline
  \rowcolor{tuluLightBlue!30}
    FAMA \texttt{medium} & \multirow{2}{*}{878M} & 11.5 & 7.0 & 5.2 & 13.9 & 7.2 & 15.9 & 8.0 & 12.3 & 0.787 & 0.821 \\
    \textcolor{tuluBlue}{\quad \textit{+ joint CTC rescoring}} & & 11.5 & 7.7 & 5.2 & 13.5 & 7.1 & 14.9 & 7.9 & 12.0 & 0.791 & 0.818 \\
    \Xhline{1.2pt}
  \end{tabular}
  \caption{\label{tab:ASRSTres}
    ASR and ST performance of FAMA models and existing SFMs as terms of comparison. The results are reported on CommonVoice (CV), Multilingual LibriSpeech (MLS), and VoxPopuli (VP) for ASR, and on CoVoST (CVST2), and FLEURS (FLRS) for ST.
    Best values are in bold.
  }
\end{table*}

\subsection{Comparison with Existing SFMs}

In Table \ref{tab:ASRSTres}, we show the results for both ASR and ST of our FAMA models and SFMs presented in Section \ref{subsec:terms-of-comparison}. For FAMA models, we provide the scores of the ASR-only model (FAMA-ASR), obtained after pre-training, and of the final ASR+ST model, as well as the results obtained through joint CTC rescoring.

Looking at the results of FAMA-ASR, we observe that the \texttt{medium} model outperforms the \texttt{small} one, with $\sim$0.8 WER improvements on average both with and without the joint CTC rescoring. Compared to 
Whisper \texttt{medium}, FAMA achieves better results with FAMA \texttt{medium} outperforming Whisper by 4.4 WER on \textit{en} and 6.4 on \textit{it} while having a similar number of model parameters. Remarkable performance is achieved by FAMA \texttt{medium} also compared to OWSM v3.1 \texttt{medium}, with improvements of up to 1.1 WER on \textit{en} and 7.3 on \textit{it}, but also compared to Whisper \texttt{large-v3}, where similar WER scores are achieved.
Instead, SeamlessM4T models, leveraging large pretrained models such as wav2vec-BERT 2.0 (which is trained on 4.5 million hours) and NLLB (which is trained on more than 43 billion sentences),  still outperform FAMA, with the \texttt{v2-large} scoring an incredibly low WER on CommonVoice also compared to a strong competitor as Whisper \texttt{large-v3}.
Looking at the ASR results of the final FAMA models, we observe that the WER remained almost unaltered compared to the ASR-only model, as already discussed in Section \ref{subsec:pre-train}. Regarding ST results, we notice that FAMA models outperform OWSM v3.1 \texttt{medium}, with an improvement of up to 0.141 COMET by FAMA \texttt{small} and 0.152 by FAMA \texttt{medium} while still struggling to achieve the performance of Whisper and SeamlessM4T. 

These mixed outcomes--competitive ASR performance even against larger non-open models but lower ST performance--demonstrate both the feasibility of building high-quality open-science SFMs and the need for initiatives dedicated to creating OS-compliant ST datasets with human references to bridge the gap with non-open models.

\subsection{Computational Time}

As an additional comparison, we evaluate the throughput of the SFMs
on a single NVIDIA A40 40GB. 
The throughput, measured in xRTF (the inverse of the real-time factor),\footnote{\url{https://github.com/NVIDIA/DeepLearningExamples/blob/master/Kaldi/SpeechRecognition/README.md\#metrics}} is calculated as the number of seconds of processed audio divided by the compute time in seconds.
The test set used for this performance evaluation is CommonVoice on both \textit{en} and \textit{it} with a total duration of, respectively, 26.9 and 26.4 hours. For each model, we report the maximum batch size possible spanning in the range 2, 4, 8, and 16, as higher values resulted in out-of-memory issues with all models. The results are reported in Table \ref{tab:speed}. 

We notice that Whisper models are the slowest ones, with an average xRTF of 12.1 for \texttt{medium} and 7.2 for \texttt{large-v3}, making them $\sim$3-6 times slower than FAMA \texttt{medium} and $\sim$5-8 than FAMA \texttt{small}. These results can be attributed to the architectural design of Whisper models that apply an $\times$2 audio subsampling compared to the commonly used $\times$4 (as in FAMA) and introduce 
a lot of padding in shorter sequences to achieve the fixed 30-second length. 
The Seamless models, despite having no extra padding (as FAMA) and a greater audio subsampling of $\times$8,
are $\sim$2 times faster than Whisper ones but still 1.5-3 times slower for, respectively, \texttt{medium} and \texttt{v2-large}, compared to FAMA \texttt{medium} and 2-4 compared to FAMA \texttt{small}, making the FAMA model family the fastest by a large margin.

\begin{table}[t]
\addtolength{\tabcolsep}{16pt}
    \centering
    \begin{tabular}{l|c|cc|c}
    \Xhline{1.2pt}
    \color{tuluBlue}
        \multirow{2}{*}{\textbf{Model}} & \color{tuluBlue}\textbf{Batch} & \multicolumn{3}{c}{\color{tuluBlue}\textbf{xRTF} ($\uparrow$)}   \\
        \cline{3-5}
        & \color{tuluBlue}\textbf{Size} & \color{tuluGray!75}\textit{en} & \color{tuluGray!75}\textit{it} & \textit{\textbf{AVG}} \\
        \hline
        Whisper \texttt{medium} & 8 & 13.3 & 10.9  & 12.1 \\
        Whisper \texttt{large-v3} & 4 & 7.9 & 6.5 & 7.2 \\
        SeamlessM4T \texttt{medium} & 2 & 28.5 & 26.2 & 27.4 \\
        SeamlessM4T \texttt{v2-large} & 2 & 13.7 & 13.3 & 13.5 \\
        \hdashline
        \rowcolor{tuluLightBlue!30}
        FAMA \texttt{small} & 16 & \textbf{57.4} & \textbf{56.0} & \textbf{56.7} \\
        \rowcolor{tuluLightBlue!30}
        FAMA \texttt{medium} & 8 & 39.5 & 41.2 & 40.4 \\
    \Xhline{1.2pt}
    \end{tabular}
    \caption{Computational time and maximum batch size for Whisper, SeamlessM4T, and FAMA models. Best values are in bold.}
    \label{tab:speed}
\end{table}



\section{Conclusions}
In this paper, we addressed the challenges posed by the closed nature of existing SFMs, such as limited accessibility to training data and codebases, by introducing FAMA, the first large-scale open-science SFM for English and Italian. Trained on over 150k hours of exclusively OS speech, FAMA ensures full transparency, with all artifacts released under OS-compliant licenses.
Additionally, we contributed a new collection of ASR and ST pseudolabels for about 16k hours of speech data, and more than 130k hours of English and Italian automatic translations. Results show that FAMA models outperform OWSM on both ASR and ST and also achieve comparable ASR results to Whisper while being up to 8 times faster.
By providing the community with fully accessible resources, FAMA bridges the gap between advances in speech technology and open science principles, enabling fair evaluation, broader participation, and inclusivity. Future work will focus on extending FAMA to additional languages with the ultimate goal of further expanding the open science ecosystem to speech technologies.

\section*{Acknowledgments}
This paper has received funding from the PNRR project FAIR - Future AI Research (PE00000013),  under the NRRP MUR program funded by the NextGenerationEU, and from the European Union's Horizon research and innovation programme under grant agreement No 101135798, project Meetween (My Personal AI Mediator for Virtual MEETings BetWEEN People).
We acknowledge CINECA for the availability of high-performance computing resources and support.

\newpage

\bibliographystyle{acl_natbib}
\bibliography{mybib}

\end{document}